\documentclass[11pt]{article}
\pdfoutput=1
\usepackage[final]{acl}

\usepackage{times}
\usepackage{latexsym}

\usepackage[T1]{fontenc}

\usepackage[utf8]{inputenc}

\usepackage{microtype}

\usepackage{inconsolata}

\usepackage[utf8]{inputenc}

\usepackage{tikz}
\usepackage{booktabs}
\usepackage{tabularx}
\usepackage{pgfplots}
\usepackage{microtype}

\usepackage{graphicx}
\usepackage{booktabs}  
\usepackage{tabularx}  
\usepackage{xcolor}    
\usepackage{amsmath}   
\usepackage{amssymb}   
\usepackage{multirow}  

\usepackage[most]{tcolorbox}

\newtcolorbox{observationbox}[1]{
    colback=gray!10,       
    colframe=gray!50,       
    arc=2pt,                
    boxrule=0.8pt,          
    left=5pt,               
    right=5pt,              
    top=5pt,                
    bottom=5pt,             
    title={\textbf{#1}},   
    coltitle=black,         
    fonttitle=\small\itshape,
    titlerule=0.5pt,        
    titlerule style=gray!30
}

\title{DiffER: Diffusion Entity-Relation Modeling for Reversal Curse \\ in Diffusion Large Language Models}

\author{
  Shaokai He$^1$, Kaiwen Wei$^1$\footnotemark[2],   
  \textbf{Xinyi Zeng}$^2$, \textbf{Xiang Chen}$^4$ \\ \textbf{Xue Yang}$^3$,\textbf{Zhenyang Li}$^1$,\textbf{Jiang Zhong}$^1$,\textbf{Yu Tian}$^2$\footnotemark[2]\\
  $^1$Chongqing University \quad $^2$Tsinghua University \quad $^3$Shanghai Jiao Tong University\\ $^4$Nanjing University of Aeronautics and Astronautics \quad $^5$Hong Kong University of Science and Technology \\
\texttt{20230271@stu.cqu.edu.cn, weikaiwen@cqu.edu.cn, tianyu1810613@gmail.com}
}

\begin{document}
\maketitle
{
  \renewcommand{\thefootnote}{\fnsymbol{footnote}} 
  

  \footnotetext[2]{Corresponding author} 
}
\begin{abstract}
The "reversal curse" refers to the phenomenon where large language models (LLMs) exhibit predominantly unidirectional behavior when processing logically bidirectional relationships. Prior work attributed this to autoregressive training—predicting the next token inherently favors left-to-right information flow over genuine bidirectional knowledge associations. However, we observe that Diffusion LLMs (DLLMs), despite being trained bidirectionally, also suffer from the reversal curse. To investigate the root causes, we conduct systematic experiments on DLLMs and identify three key reasons: 1) entity fragmentation during training, 2) data asymmetry, and 3) missing entity relations. Motivated by the analysis of these reasons, we propose \textbf{Diff}usion \textbf{E}ntity-\textbf{R}elation Modeling (\textbf{DiffER}), which addresses the reversal curse  through entity-aware training and balanced data construction. Specifically, DiffER introduces whole-entity masking, which mitigates entity fragmentation by predicting complete entities in a single step. DiffER further employs distribution-symmetric and relation-enhanced data construction strategies to alleviate data asymmetry and missing relations. Extensive experiments demonstrate that DiffER effectively alleviates the reversal curse in Diffusion LLMs, offering new perspectives for future research.

\end{abstract}

\section{Introduction}

With the continuous development of Large language models (LLMs), their applications expand across various industries. Some work~\cite{berglund2023reversal,ma2023untying,allen2023physics,zhu2024towards} finds that current LLMs face a significant challenge due to the "reversal curse." This curse describes a phenomenon where LLMs exhibit unidirectional behavior with symmetric bidirectional relationships. For example, after learning from the "X's father is Y," LLMs can answer "Who is X's father?" correctly, but struggle with "Who is Y's child?". This limitation severely impacts the reasoning capabilities and knowledge generalization of LLMs, highlighting a fundamental gap between pattern matching and relational understanding. Therefore, addressing the reversal curse is essential for creating more intelligent and reliable LLMs.

\begin{figure}[t]
  \includegraphics[width=\columnwidth]{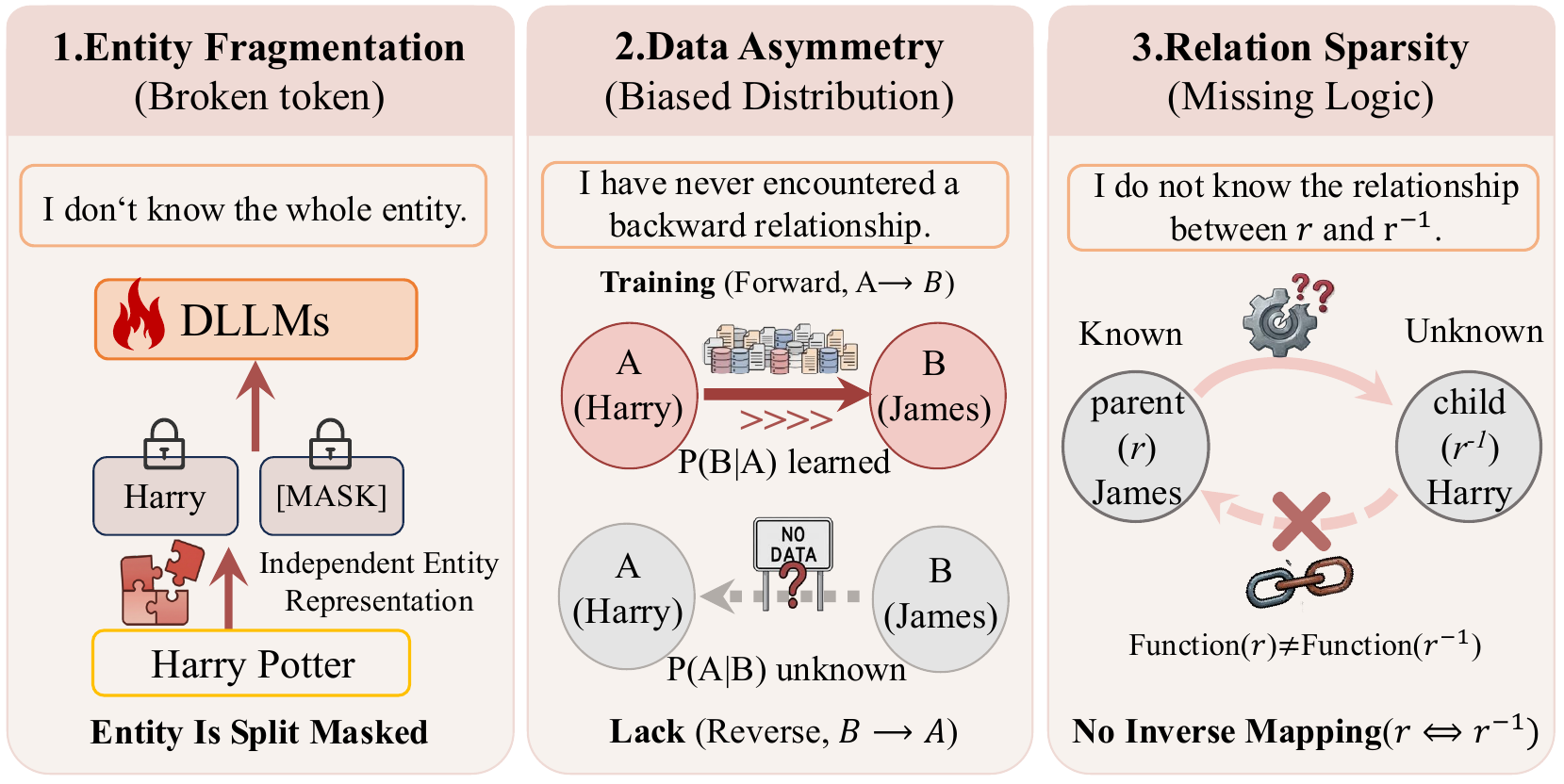}
  \caption{The identified 3 factors of DLLM reversal curse: 1) entity fragmentation (split representation), 2) data asymmetry (directional bias), and 3) relationship sparsity (missing logic).}
  \label{fig:category}
\end{figure}

Previous research~\cite{berglund2023reversal,allen2023physics,zhu2024towards} primarily attributed the reversal curse to the autoregressive training mechanism of LLMs, where the model predicts the next token sequentially from left to right. This unidirectional training objective leads to a sequential processing of information, making it challenging to establish true bidirectional knowledge associations.  With the emergence of Diffusion LLMs (DLLMs), many researchers~\cite{pan2025closing} believed that this issue had been resolved. However, our pilot experiments indicate that even DLLMs using bidirectional architectures remain impacted by the reversal curse. For example, as shown in Table~\ref{tab:factors_detailed}, while the model achieves 92.00\% accuracy on forward queries (e.g., ``Who is A's parent?''), its performance collapses to approximately 46.73\% on reverse queries (e.g., ``Who is B's child?''), often resulting in fragmented or hallucinatory outputs.  This situation prompts us to explore the underlying reasons for this persistence, which could provide valuable insights for both DLLMs and LLMs.


To systematically explore and quantify the mechanisms behind the reversal curse, we conduct a series of cause analysis experiments on DLLMs, evaluating their performance under two-stage training  protocol on the PORE dataset \citep{lu2024rethinking}. Through fine-grained analysis, as shown in Fig.~\ref{fig:category}, we identify 3 main factors that contribute to this issue: 1) \textbf{entity fragmentation}, where the complete entity is split across multiple prediction steps, disrupting consistency in its representation; 2) \textbf{data asymmetry}, which leads to bias in model learning due to imbalances in the distribution of forward and backward relationship; and 3) \textbf{relationship sparsity}, where the explicit encoding of entity relationships in the training data is insufficient.








Inspired by these findings, we design \textbf{Diff}usion \textbf{E}ntity-\textbf{R}elation Modeling (\textbf{DiffER}), a post-training method that addresses the reversal curse in DLLMs through entity-aware training and balanced data construction. First, inspired by the whole-word masking mechanism~\cite{cui2021pre}, we propose a full entity masking training strategy to address entity fragmentation by ensuring that entire entities are predicted within a single denoising step, thus preserving their integrity during the learning process. Then, we introduce a data construction strategy that enhances symmetry and relational relevance, mitigating data asymmetry and relationship sparsity through the creation of reversed entity data and relationship prediction data. Extensive experiments demonstrate that the proposed DiffER method significantly reduces the reversal curse in DLLMs, leading to substantial improvements in bidirectional reasoning tasks. In summary, the main contributions of this paper include:

(1) We demonstrate that Diffusion Large Language Models still suffer from the reversal curse and systematically identify entity fragmentation, data asymmetry, and relation sparsity as the three primary root causes.

(2) We introduce DiffER to address the reversal curse through three key components: whole-entity masking, symmetric alignment, and inverse relation modeling.

(3) Extensive experiments on the parent–child and company-ceo datasets validate that DiffER could effectively mitigate the reversal curse. 

\section{Related Work}

\subsection{Reversal Curse}
The "Reversal Curse" denotes the inability of LLMs to generalize from $A \to B$ to $B \to A$, highlighting fundamental deficits in factual consistency and knowledge editing \citep{berglund2023reversal,elazar2021measuring, cohen2024evaluating, lu2024rethinking}.
While data augmentation can mitigate specific instances, it lacks robustness across unseen relations or complex compositions \citep{grosse2023studying}.
Mechanistically, this asymmetry stems from the autoregressive objective: Transformer FFNs function as unidirectional Key-Value memories where value access does not trigger keys \citep{geva2021transformer, dai2022knowledge, meng2022locating}, encoding entities as surface-level dependencies \citep{allen2023physics}.
However, these explanations rely almost exclusively on autoregressive architectures, leaving unclear whether the Reversal Curse reflects an inherent limitation of relational generalization or merely an artifact of autoregressive training.

\subsection{Diffusion LLMs}
Diffusion LLMs (DLLMs) represent a paradigm shift towards non-autoregressive generation, iteratively denoising sequences via bidirectional attention mechanisms \citep{li2022diffusion,he2023diffusionbert,gong2022diffuseq,nie2025large,ye2025dream}. Unlike causal decoders constrained by unidirectional masking, DLLMs leverage omnidirectional context throughout the learning process.
This field has rapidly evolved from early discrete corruption frameworks \citep{austin2021structured} to scalable architectures like SEDD \citep{lou2023discrete}, MDLM \citep{sahoo2024simple}, and simplified diffusion objectives \citep{shi2024simplified}. Theoretically, DLLMs' global visibility suggests potential immunity to the Reversal Curse \citep{pan2025closing}. However, overcoming it requires explicit alignment in training strategy and data construction, not architecture alone.


\begin{table*}[t!]
    \centering
    \footnotesize
    
    
    \begin{minipage}[c]{0.38\textwidth}
        \centering
        \begin{tikzpicture}[baseline={([yshift={-0.8em}]current bounding box.center)}]
            \begin{axis}[
                ybar,
                width=\linewidth,
                height=4.5cm,
                enlarge x limits=0.5,
                ymin=0, ymax=65,
                ylabel={Error Rate (\%)},
                ylabel style={font=\footnotesize, yshift=-1em},
                xtick={1, 2},
                xticklabels={Total\\Error, Entity\\Fragmentation},
                xticklabel style={font=\footnotesize, align=center, text width=2cm},
                nodes near coords,
                nodes near coords style={font=\footnotesize\bfseries, color=black},
                axis x line*=bottom,
                axis y line*=left,
                ymajorgrids=true,
                grid style={dashed, gray!30},
                bar width=1.1cm,
                ytick={0,20,40,60},
                bar shift=0pt,
            ]
                \addplot+[fill=gray!30, draw=none] coordinates {(1, 53.27)};
                \addplot+[fill=orange!60, draw=none] coordinates {(2, 26.97)};
            \end{axis}
        \end{tikzpicture}
    \end{minipage}%
    \hfill
    \begin{minipage}[c]{0.60\textwidth}
        \centering
        \setlength{\tabcolsep}{4pt}
        \begin{tabularx}{\linewidth}{l >{\centering\arraybackslash}X c c}
            \toprule
            \textbf{Factors} & \textbf{Inference Query} & \textbf{Acc (\%)} & \textbf{Gap ($\Delta$)} \\
            \midrule
            \textbf{Baseline}
            & Who is A's parent? ($A \to B$)
            & 92.00 & - \\
            \addlinespace[0.8em]
            \textbf{Relation Sparsity}
            & Whose child is A? ($A \to B$)  
            & 24.45 & \textcolor{red}{-67.55} \\
            \addlinespace[0.8em]
            \textbf{Data Asymmetry}
            & Whose parent is B? ($B \to A$) 
            & 24.92 & \textcolor{red}{-67.08} \\
            \bottomrule
        \end{tabularx}
    \end{minipage}
    
    \par 
    
    \begin{minipage}[t]{0.38\textwidth}
        \centering 
        (a) Entity fragmentation problem in DLLM
    \end{minipage}%
    \hfill
    \begin{minipage}[t]{0.60\textwidth}
        \centering 
        (b) Relation sparsity and data asymmetry problems in DLLM
    \end{minipage}

    \caption{
    Pilot analysis of reversal curse causes: (a) Entity Fragmentation: 26.97\% of errors are due to fragmented representations of multi-token entities.(b) Generalization gaps: Although baseline accuracy reaches 92\%, performance degrades under relation sparsity (failing on paraphrased queries) and data asymmetry (reverse $B\rightarrow A$).
    }
    \label{tab:factors_detailed}
\end{table*}

\section{Pilot Experiments}
\label{sec:preliminary}
To systematically investigate whether DLLMs inherently mitigate the "reversal curse," we conduct a series of controlled pilot experiments. Specifically, we evaluate DLLM performance under two distinct training paradigms using a controlled reversal dataset. Our fine-grained analysis identifies three primary factors: 1) \textit{Entity Fragmentation}, 2) \textit{Data Asymmetry}, and 3) \textit{Relation Sparsity}.


\subsection{Experimental Setup}
We design a two-stage protocol to disentangle the effects of knowledge acquisition from format alignment. To ensure rigorous benchmarking, we strictly align our data construction for both knowledge injection and subsequent inference with the PORE dataset specifications \citep{lu2024rethinking}.



\paragraph{Stage 1: Knowledge Injection via Continued Pre-training.}
We perform continued pre-training on \text{LLaDA-8B-Base}~\citep{nie2025large} using the parent--child subset from PORE benchmark. 
The base corpus $D_{\mathrm{base}}$ contains only forward declarative statements (e.g., ``A's parent is B''); all reverse patterns and paraphrases are filtered to prevent information leakage. We optimize using LLaDA's discrete denoising objective, which reconstructs randomly masked tokens . Crucially, while masking is bidirectional, entity order remains fixed ($A \rightarrow B$). This tests whether symmetric generalization can emerge solely from bidirectional architecture despite unidirectional structural exposure.

\paragraph{Stage 2: Instruction Alignment via Prompt-Conditioned SFT.}
To evaluate retrieved knowledge, we align the pre-trained model to follow Question-Answering instructions. We construct prompt--response pairs derived strictly from the pre-training facts (e.g., Prompt: \textit{``A's parent is whom?''}, Response: $B$).
We adapt LLaDA's denoising mechanism by keeping the Prompt ($x_{\mathrm{p}}$) unmasked while diffusing the Response($x_{\mathrm{r}}$). This forces the model to utilize the knowledge injected in Stage 1 to generate $B$ conditioned on $A$, explicitly modeling the conditional probability $P(B|A)$.

\begin{figure*}[t] 
    \centering
    \includegraphics[width=1 \textwidth]{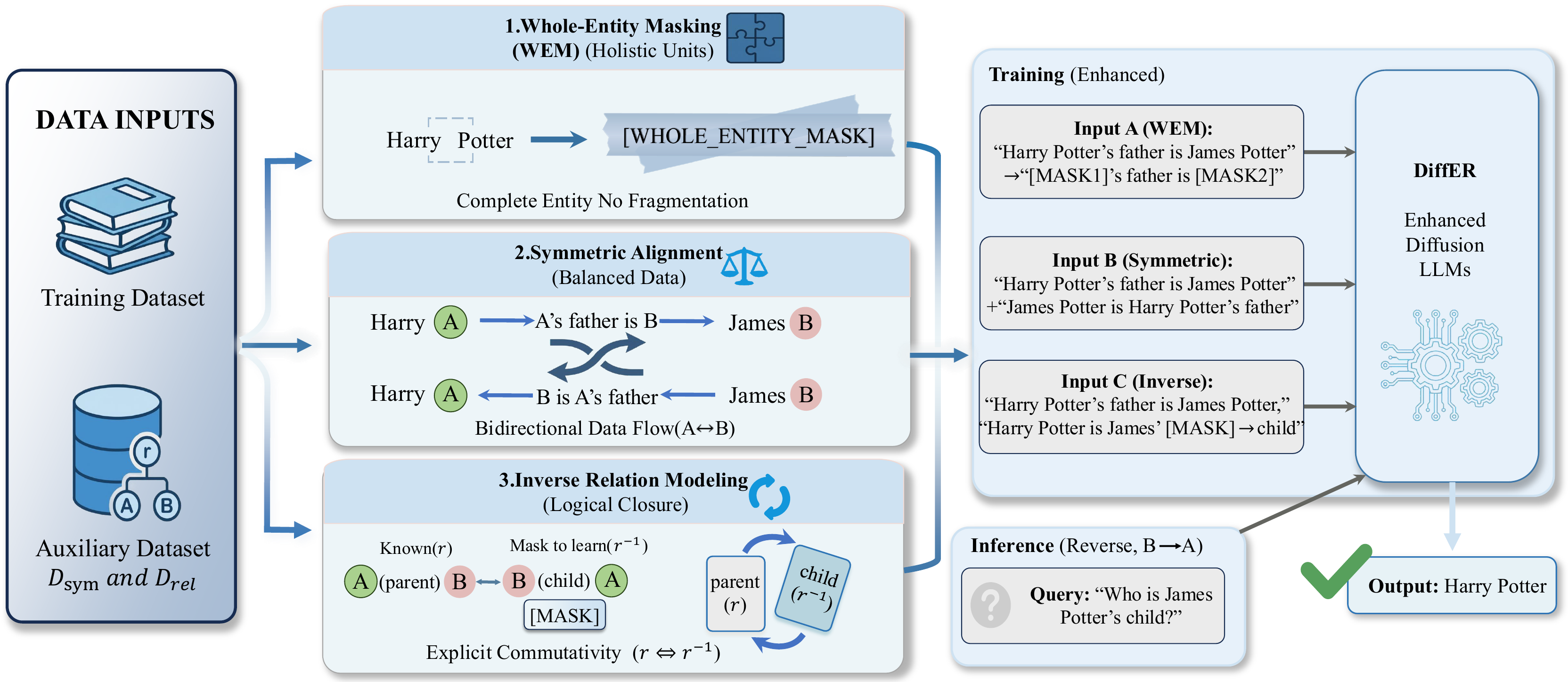} 
    \caption{
    Overview of DiffER. (1) WEM: entity-level denoising for structural integrity; (2) Symmetric Alignment: bidirectional balancing via distribution-symmetric $D_{\text{sym}}$ augmentation; (3) Inverse Relation Modeling: logical commutativity via relation prediction on relational dataset $D_{\text{rel}}$.
    }
    \label{fig:method}
\end{figure*}

\subsection{Verification and Root Cause Analysis}
To verify the causes of the reversal curse in DLLMs, we evaluate the model from three dimensions:

\paragraph{Evaluation on Qualitative Errors.} 
 As shown in Table~\ref{tab:factors_detailed} (a), failures in \textit{Logical Compositionality} (e.g., deducing \textit{``Who is B's child?''}) often manifest as partial hallucinations. This stems from the model treating multi-token entities (e.g., ``Harry Potter'') as disjoint sequences (e.g., ``Harry'' and ``Potter'') rather than atomic concepts. We term this \textbf{Entity Fragmentation}, where the integrity of multi-token entities is compromised during the denoising process.

\paragraph{Evaluation on Forward Consistency ($A \to B$).} 
We assess the model's understanding of the trained direction by comparing performance on \textit{Knowledge Retention} queries (identical to training patterns, e.g., \textit{``Who is A's parent?''}) with \textit{Semantic Paraphrasing} queries (using inverse phrasing, e.g., \textit{``Whose child is A?''}). As shown in Table \ref{tab:factors_detailed} (b), while the model exhibits near-perfect recall on exact training patterns, it suffers a significant performance drop when the phrasing is altered. This discrepancy suggests an inability to internalize the underlying semantic relationship, pointing to \textbf{Relation sparsity}, where the model lacks the ability to compute commutative mappings.

\paragraph{Evaluation on Backward Generalization ($B \to A$).} 
We probe the model's ability to generalize beyond the training flow, including \textit{Reverse Inference} queries (inferring the subject from the object, e.g., \textit{``Whose parent is B?''}) . As shown in Table \ref{tab:factors_detailed} (b), the model experiences a catastrophic collapse on direct reversal tasks, confirming a profound \textbf{Data Asymmetry} in conditional modeling where $P(A|B) \ll P(B|A)$.

\paragraph{Motivation.} 
Our empirical analysis reveals 3 distinct causes: (1) \textbf{Entity Fragmentation}, evidenced by partial hallucinations due to disjoint token processing; (2) \textbf{Data Asymmetry}, evidenced by the inability to reverse direction; and (3) \textbf{Relation Sparsity},  evidenced by the brittleness to phrasing changes. Motivated by this, we propose \textbf{DiffER}, which alleviates reversal curse by entity-aware training and balanced data construction.

\section{Methodology}
\label{sec:methodology}

Inspired by the finding from pilot experiments, we propose DiffER, a post-training method designed to mitigate the reversal curse in DLLMs. As illustrated in Figure~\ref{fig:method}, DiffER addresses identified bottlenecks via 3 integrated components: (1) \textbf{Whole-Entity Masking (WEM)}, which counters entity fragmentation by preserving the structural integrity of multi-token concepts; (2) \textbf{Symmetric Alignment}, which mitigates data asymmetry by rebalancing directional probability distributions; and (3) \textbf{Inverse Relation Modeling}, which resolves relation sparsity by explicitly encoding logical commutativity during denoising.



\subsection{Whole-Entity Masking}
\label{ssec:wem}

A critical observation from the pilot experiments is that independent token-level masking often compromises the semantic cohesion of multi-token entities, leading to fragmented representations. To rectify this, we propose \textbf{Whole-Entity Masking (WEM)} mechanism during pre-training, which transitions from standard stochastic noise schedules to a structure-aware corruption process. In this regime, entities are treated as atomic semantic units rather than disjoint sub-word sequences.

Specifically, given an input sequence $\mathbf{x} = (t_1, \dots, t_n)$ and a set of entity spans $\mathcal{S} = \{(i,j)\}$, where each $(i,j)$ corresponds to a contiguous token span representing a named entity, we first obtain a provisional token-level mask $\tilde{M} \in \{0,1\}^n$ following the base corruption process of the DLLM~\cite{nie2025large}. WEM then applies an \emph{entity-level contagion rule}. Specifically, for any entity  spanning indices $(i, j)$ in the sequence, if any token within this span is selected by the base mask (i.e., partial occlusion), the mask is propagated to cover the entire span $(i, j)$. To resolve potential conflicts from nested entities (e.g., "New York" inside "New York City"), we employ a longest-match-first heuristic, prioritizing the preservation of maximal semantic units during the boundary alignment phase to construct a structure-aware mask $M$ such that, for each entity span $(i,j) \in \mathcal{S}$,
\begin{equation}
\small
M_k =
\begin{cases}
1, & \text{if } \exists\, m \in [i,j] \text{ s.t. } \tilde{M}_m = 1 \text{ and } k \in [i,j], \\
\tilde{M}_k, & \text{otherwise}.
\end{cases}
\end{equation}

This monolithic masking constraint ensures that entity mentions are corrupted in an all-or-nothing manner. For example, consider the entity ``\textit{New York}''. Under standard token-level masking, it is possible for only ``\textit{New}'' to be masked while ``\textit{York}'' remains visible, enabling the model to trivially reconstruct the masked token through shallow intra-entity cues. Such partial exposure encourages memorization of surface forms rather than abstraction of the entity as a unified concept. In contrast, WEM enforces that once any constituent token of an entity is selected for corruption, the entire entity span is masked simultaneously. Consequently, the model is forced to reconstruct the entity holistically using global contextual information, rather than relying on local token adjacency. 
The corresponding training objective for WEM is defined as an entity-aligned denoising loss:
\begin{equation}
\small
\mathcal{L}_{\text{wem}} =
\mathbb{E}_{(\mathbf{x}, M)}
\Big[
- \sum_{k=1}^{n} M_k \log P_\theta(t_k \mid \mathbf{x}_{\setminus M})
\Big],
\end{equation}
where $M$ is the structure-aware mask constructed by WEM, $\mathbf{x}_{\setminus M}$ denotes the corrupted input sequence, and $P_\theta$ is the model's conditional token distribution.

By eliminating partial entity exposure, WEM directly mitigates the {entity fragmentation} phenomenon identified in our pilot study. The model is trained to minimize the cross-entropy loss over tokens selected by $M$, thereby enforcing an indivisible denoising objective at the entity level. This inductive bias promotes coherent entity representations and forms a critical foundation for alleviating downstream failures associated with reversal and compositional generalization.

\subsection{Symmetric Alignment}
\label{ssec:symmetric_alignment}

DLLMs may still inherit the data asymmetry problem induced by large-scale pre-training, where forward conditional dependencies dominate their reverse counterparts. In particular, for a factual relation $(A, r, B)$, the learned conditional distribution often satisfies $P(B \mid A) \gg P(A \mid B)$, resulting in a skewed probability landscape that hinders backward inference.

To correct this imbalance, we introduce a \textbf{Symmetric Alignment} strategy during the SFT stage. The core of this strategy is the construction of a {distribution-symmetric dataset} $D_{\text{sym}}$, which explicitly counteracts forward-only inductive biases. For each factual triple $(A, r, B)$ in the original corpus, we augment the training data with its logically reversed declarative form $(B, r, A)$, together with corresponding question--answering pairs. For example, a forward statement such as ``\textit{A's parent is B}'' is paired with its inverse ``\textit{B is A's parent}.''

Training on $D_{\text{sym}}$ encourages the model to approximate the joint distribution $P(A, B)$ rather than overfitting to a single conditional direction. Concretely, the SFT objective can be viewed as minimizing the follow equation:
\begin{equation}
\small
\mathcal{L}_{\text{sym}} = 
\mathbb{E}_{(A,B)\sim D_{\text{sym}}}
\big[ - \log P_\theta(B \mid A) - \log P_\theta(A \mid B) \big],
\end{equation}
which thereby enforcing statistical parity between forward and backward inference paths. As a result, the model mitigates directional bias and achieves bidirectional accessibility of stored relational knowledge.




\subsection{Inverse Relation Modeling}
\label{ssec:inverse_modeling}

As revealed by our analysis of {relation sparsity}, models often fail to infer an inverse relation $r^{-1}$ (e.g., \textit{child}) even when the corresponding base relation $r$ (e.g., \textit{parent}) is well memorized. This failure indicates that inverse relations are not implicitly encoded as logical transformations within the latent space, but are instead treated as independent surface patterns.

To address this limitation, we introduce an \textbf{Inverse Relation Modeling} objective supported by a dedicated {relational dataset} $D_{\text{rel}}$ during the SFT stage. Unlike entity-centric masking strategies, this objective explicitly targets the {relational descriptor}, encouraging the model to learn logical entailment between paired relations.

Concretely, we construct {logical implication chains} that explicitly connect forward facts to their inverse counterparts. Each training instance is formulated as a deductive sequence, such as: ``\textit{$A$'s parent is $B$. Therefore, $A$ is $B$'s [MASK].}'' Given the co-occurrence of entities $A$ and $B$ and the observed relation $r$, the model is required to reconstruct the masked inverse relation token $r^{-1}$. This learning objective can be expressed as minimizing: 
\begin{equation}
\small
\mathcal{L}_{\text{rel}} = 
\mathbb{E}_{(A,B,r,r^{-1}) \sim D_{\text{rel}}}
\big[ - \log P_\theta(r^{-1} \mid A, B, r) \big],
\end{equation}
which explicitly enforces the conditional mapping from $(A, B, r)$ to its inverse relation $r^{-1}$.

By shifting the denoising focus from entity reconstruction to relational inference, this objective encourages the model to internalize the \emph{semantic reciprocity} between paired relations. As a result, the model moves beyond memorization of isolated relational patterns and acquires the ability to compute bidirectional relational mappings, thereby resolving relation sparsity at a structural level.

\begin{table*}[t!]
\centering

\renewcommand{\arraystretch}{1.2} 

\setlength{\abovecaptionskip}{10pt plus 3pt minus 2pt}
\setlength{\belowcaptionskip}{0pt}

\resizebox{\textwidth}{!}{%
\begin{tabular}{l | cccc @{\hspace{0.8cm}} | cccc}
\toprule
\multirow{2.5}{*}{\textbf{Training Instruction}} & \multicolumn{4}{c|}{\textbf{Standard LLaDA (Baseline)}} & \multicolumn{4}{c}{\textbf{ DiffER (Ours)}} \\
\cmidrule(lr){2-5} \cmidrule(lr){6-9}
 & \shortstack{\textit{Who is A's}\\\textit{parent?}} & \shortstack{\textit{Whose child}\\\textit{is A?}} & \shortstack{\textit{Whose parent}\\\textit{is B?}} & \shortstack{\textit{Who is B's}\\\textit{child?}} & \shortstack{\textit{Who is A's}\\\textit{parent?}} & \shortstack{\textit{Whose child}\\\textit{is A?}} & \shortstack{\textit{Whose parent}\\\textit{is B?}} & \shortstack{\textit{Who is B's}\\\textit{child?}} \\
\midrule
\textit{Train:A's parent is whom?B} & \textbf{92.00} & 24.45 & 24.92 & 46.73 & \textbf{97.75} & 28.08 & 26.31 & 49.83 \\
\addlinespace[0.4em]
\textit{Train:A is whose child?B} & 6.15 & \textbf{83.28} & 13.28 & 0.26 & 12.56 & \textbf{87.24} & 15.47 & 1.85 \\
\addlinespace[0.4em]
\textit{Train:B is Whose parent?A} & 10.44 & 14.14 & \textbf{82.68} & 19.76 & 12.89 & 17.38 & \textbf{89.56} & 22.47 \\
\addlinespace[0.4em]
\textit{Train:B's child is whom?A} & 47.85 & 0.99 & 12.03 & \textbf{94.84} & 50.89 & 1.65 & 18.18 & \textbf{97.88} \\
\bottomrule
\end{tabular}%
}
\caption{
Performance comparison (\%) on the parent–child benchmark between Standard LLaDA (left) and DiffER (right) across 4 instruction templates. Diagonal values denote in-template evaluation, while off-diagonal values reflect generalization to alternative phrasings or directions.}
\label{tab:comprehensive_matrix}
\end{table*}

\section{Experiments}
\subsection{Experimental Setup}

\paragraph{Datasets.} 
To systematically evaluate the efficacy of the proposed DiffER framework, we conduct extensive experiments on two datasets from the PORE benchmark \citep{lu2024rethinking}: {parent-child} and {company-ceo}. For the {parent-child} benchmark, we leverage the entire dataset to construct the base pre-training dataset $\mathcal{D}_{\text{base}}$ containing 1,513 high-quality factual pairs, all instantiated as forward declarative statements. To facilitate symmetric alignment and inverse relation modeling, we further derive two auxiliary subsets by randomly sampling from $\mathcal{D}_{\text{base}}$: a symmetric alignment corpus $\mathcal{D}_{\text{sym}}$ (200 pairs) comprising both forward and inverse formulations, and a relational inference corpus $\mathcal{D}_{\text{rel}}$ (200 pairs) specifically designed to supervise inverse relation prediction. To assess the generalizability of DiffER, we extend this setup to the {company-ceo} benchmark, following an identical construction pipeline with $\mathcal{D}_{\text{base}}$, $\mathcal{D}_{\text{sym}}$, and $\mathcal{D}_{\text{rel}}$ sizes of 1,697, 200, and 200, respectively.

\paragraph{Evaluation.} Following previous research~\cite{lu2024rethinking}, we adopt exact-match accuracy as the primary metric. Specifically, it calculates the proportion of instances where the model's output string matches the reference answer exactly. It serves as a rigorous indicator of the model's precision in generating factual and well-defined responses.


\noindent \textbf{Implementation Details.}
We employ \texttt{LLaDA-8B} \citep{nie2025large} and \texttt{Dream} \citep{ye2025dream} as the foundational backbones. For the pre-training phase, we set the maximum sequence length to 4,096 tokens to facilitate long-context knowledge acquisition, training for 3 epochs with a learning rate of $1 \times 10^{-5}$. In the SFT phase, we truncate the sequence length to 128 tokens to align with the concise nature of the datasets. This phase lasts for 50 epochs to ensure convergence, maintaining a learning rate of $1 \times 10^{-5}$. Across all stages, we utilize the AdamW optimizer with a linear warmup ratio of 0.03. All experiments are conducted on two NVIDIA RTX A800 GPUs.


\begin{table*}[t!]
\centering

\renewcommand{\arraystretch}{1.2} 

\setlength{\abovecaptionskip}{10pt plus 3pt minus 2pt}
\setlength{\belowcaptionskip}{0pt}

\resizebox{\textwidth}{!}{%
\begin{tabular}{l | cccc | cccc | cccc}
\toprule
\multirow{2.5}{*}{Training Instruction}
& \multicolumn{4}{c|}{Whole-Entity Masking}
& \multicolumn{4}{c|}{Symmetric Alignment}
& \multicolumn{4}{c}{Inverse Modeling} \\
\cmidrule(lr){2-5}
\cmidrule(lr){6-9}
\cmidrule(lr){10-13}
& \shortstack{\textit{Who is}\\\textit{A's parent?}}
& \shortstack{\textit{Whose child}\\\textit{is A?}}
& \shortstack{\textit{Whose parent}\\\textit{is B?}}
& \shortstack{\textit{Who is}\\\textit{B's child?}}
& \shortstack{\textit{Who is}\\\textit{A's parent?}}
& \shortstack{\textit{Whose child}\\\textit{is A?}}
& \shortstack{\textit{Whose parent}\\\textit{is B?}}
& \shortstack{\textit{Who is}\\\textit{B's child?}}
& \shortstack{\textit{Who is}\\\textit{A's parent?}}
& \shortstack{\textit{Whose child}\\\textit{is A?}}
& \shortstack{\textit{Whose parent}\\\textit{is B?}}
& \shortstack{\textit{Who is}\\\textit{B's child?}} \\
\midrule
\textit{Train:A's parent is whom?B}
& 96.23 & 26.57 & 25.18 & 48.84
& 95.44 & 26.37 & 25.45 & 47.79
& 95.77 & 27.03 & 24.98 & 49.17 \\
\addlinespace[0.4em]
\textit{Train:A is whose child?B}
& 11.70 & 85.86 & 14.61 & 1.19
& 11.57 & 83.74 & 13.95 & 0.66
& 12.10 & 84.53 & 15.27 & 0.79 \\
\addlinespace[0.4em]
\textit{Train:B is Whose parent?A}
& 11.70 & 15.66 & 87.57 & 21.08
& 11.96 & 15.07 & 88.57 & 20.95
& 11.57 & 14.67 & 86.65 & 21.35 \\
\addlinespace[0.4em]
\textit{Train:B's child is whom?A}
& 50.36 & 1.19 & 17.58 & 98.02
& 48.78 & 1.39 & 17.32 & 96.03
& 49.44 & 1.06 & 17.85 & 97.62 \\
\bottomrule
\end{tabular}
}
\caption{
Ablation study of different components of DiffER on the parent–child benchmark.
}
\label{tab:ablation_right_only}
\end{table*}

\begin{table*}[t!]
\centering

\renewcommand{\arraystretch}{1.2} 

\setlength{\abovecaptionskip}{10pt plus 3pt minus 2pt}
\setlength{\belowcaptionskip}{0pt}

\resizebox{\textwidth}{!}{%
\begin{tabular}{l | cccc @{\hspace{0.8cm}} | cccc}
\toprule
\multirow{2.5}{*}{\textbf{Training Instruction}} & \multicolumn{4}{c|}{\textbf{Standard LLaDA (Baseline)}} & \multicolumn{4}{c}{\textbf{ DiffER (Ours)}} \\
\cmidrule(lr){2-5} \cmidrule(lr){6-9}
 & \shortstack{\textit{Who is A's}\\\textit{ceo?}} & \shortstack{\textit{Whose company}\\\textit{is A?}} & \shortstack{\textit{What is B }\\\textit{ceo of?}} & \shortstack{\textit{What is B's}\\\textit{company?}} & \shortstack{\textit{Who is A's}\\\textit{ceo?}} & \shortstack{\textit{Whose company}\\\textit{is A?}} & \shortstack{\textit{what is B}\\\textit{ceo of?}} & \shortstack{\textit{What is B's}\\\textit{company?}} \\
\midrule
\textit{Train:A's ceo is whom?B} & \textbf{83.74} & 77.78 & 0.35 & 0.29 & \textbf{90.87} & 83.21 & 2.71 & 2.89 \\
\addlinespace[0.4em]
\textit{Train:A is whose company?B} & 80.26 & \textbf{81.79} & 0.18 & 0.24 & 86.15 & \textbf{90.45} & 3.07 & 2.47 \\
\addlinespace[0.4em]
\textit{Train:B is ceo of what?A} & 0.24 & 0.12 & \textbf{88.04} & 83.97 & 2.77 & 3.12 & \textbf{95.17} & 87.06 \\
\addlinespace[0.4em]
\textit{Train:B's company is what?A} & 0.18 & 0.24 & 63.23 & \textbf{89.33} & 3.48 & 2.95 & 69.18 & \textbf{96.52} \\
\bottomrule
\end{tabular}%
}
\caption{
Performance comparison (\%) on the company-ceo benchmark. }
\label{tab:company_ceo_result}
\end{table*}
\subsection{Main Experiment}
Table~\ref{tab:comprehensive_matrix} summarizes the performance of the baseline LLaDA and DiffER under different training–evaluation instruction combinations on the parent–child benchmark. Overall, DiffER consistently preserves or improves performance on evaluation templates that match the training instruction (diagonal entries). For example, under “Who is A’s parent?”, DiffER improves accuracy from 92.00\% to 97.75\%. This indicates that the proposed modifications do not impair factual memorization, but instead lead to more precise and stable entity-level denoising during retrieval.

More importantly, DiffER exhibits stronger generalization when the evaluation instruction deviates from the training direction or surface form. While the baseline LLaDA suffers a substantial performance drop when inference shifts from the trained $A \rightarrow B$ direction to reversed $B \rightarrow A$ queries, DiffER consistently maintains a higher accuracy ceiling. In particular, on the reversed template “Who is B’s child?”, DiffER achieves an absolute improvement from 46.73\% to 49.83\%. Beyond simple reversal, DiffER also improves robustness to semantic paraphrasing, with consistent gains observed across alternative formulations. These results suggest that DiffER captures more coherent entity–relation associations, enabling diffusion-based LLMs to generalize beyond instruction-specific patterns and better support bidirectional relational retrieval.

\subsection{Ablation Study}

To quantify the contribution of each component in DiffER, we conduct an ablation study by independently enabling Whole-Entity Masking (WEM), Symmetric Alignment, and Inverse Modeling. Results are reported in Table~\ref{tab:ablation_right_only}. 

\textbf{Component-wise Effects.} Each component addresses a distinct failure mode. WEM primarily improves robustness to entity fragmentation, yielding strong performance on forward and paraphrased queries (e.g., 85.86\% on ``Whose child is A?'' when trained on the same template). However, its impact on reverse-direction inference remains limited, indicating that improved entity integrity alone is insufficient to correct directional bias.

In contrast, Symmetric Alignment and Inverse Modeling substantially improve reverse inference and logical compositionality. By rebalancing directional supervision and explicitly modeling reciprocal relations, these components enable effective backward reasoning. For example, with Inverse Modeling alone, the model achieves 49.17\% accuracy on ``Who is B’s child?'' despite being trained only on ``Who is A’s parent?'' This suggests that data-level relational supervision is the primary driver for mitigating directional asymmetry.

\textbf{Orthogonality and Synergy.}
The ablation results also indicate that the components operate on complementary dimensions. Whole-Entity Masking stabilizes entity representations, whereas Symmetric Alignment and Inverse Modeling reshape the bidirectional relational structure. Their combination yields consistent additive gains, explaining the strongest performance of the full DiffER model.

\subsection{Generalizability of DiffER}
\label{ssec:generalizability_dream}

To assess whether DiffER generalizes beyond a specific backbone, we evaluate it on Dream~\citep{ye2025dream}. We apply the same DiffER training pipeline to the \texttt{Dream-7B} backbone, comparing a standard Dream model trained on $D_{\mathrm{base}}$ with its DiffER-enhanced counterpart under identical hyperparameters.
As shown in Table~\ref{tab:dream_generalization}, the results closely mirror those observed on LLaDA. The standard Dream model suffers a substantial performance drop on reverse-direction queries, indicating that the reversal curse persists across diffusion-based architectures. In contrast, DiffER consistently improves reverse inference and relational reasoning, demonstrating that its benefits are architecture-agnostic and stem from addressing reversal curse of the diffusion objective. 


We further evaluate DiffER on the company--ceo benchmark to assess its robustness across relational domains. As shown in Table~\ref{tab:company_ceo_result}, while the baseline maintains high proficiency in standard relational queries, it exhibits severe performance degradation when handling reverse queries and combinational logic. We attribute this significant performance disparity to the inherent imbalanced data distribution within the domain. Despite this, DiffER consistently mitigates the reversal curse, mirroring the systematic improvements observed in the parent--child task. These results indicate that the gains of DiffER are not specific to a particular relation type but generalize effectively across structurally similar relational settings, demonstrating robust bidirectional reasoning.

\begin{table}[t!]
\centering
\vspace{0.2cm}
\resizebox{0.48\textwidth}{!}{%
\begin{tabular}{l | cc | cc}
\toprule
\multirow{2}{*}{\textbf{Task Metric}} & \multicolumn{2}{c|}{\textbf{Standard Dream}} & \multicolumn{2}{c}{\textbf{Enhanced Dream (Ours)}} \\
\cmidrule(lr){2-3} \cmidrule(lr){4-5}
 & \textbf{Acc (\%)} & \textbf{Gap} & \textbf{Acc (\%)} & \textbf{Gap} \\
\midrule
\textbf{Forward Retention} ($A \to B$) & 78.06 & - & 82.35 & \textcolor{teal}{+4.29}\\
\textbf{Reverse Inference} ($B \to A$) & 21.28 & \textcolor{red}{-56.78} & 23.73 & \textcolor{teal}{+2.45} \\
\textbf{Logical Reasoning} ($r \to r^{-1}$) & 21.81 & \textcolor{red}{-56.25} & 24.39 & \textcolor{teal}{2.58} \\
\bottomrule
\end{tabular}%
}
\caption{Comparison between the Standard Dream baseline and our DiffER model. }
\label{tab:dream_generalization}
\end{table}

\subsection{Error Analysis}
To further investigate how DiffER mitigates specific failure modes, we conduct a detailed error analysis on all incorrect predictions in the parent--child benchmark for the {``Who is B’s child?''} inference query when trained with the {``A’s parent is B''} template. We utilize error rate as evaluation criteria and categorize erroneous outputs into 3 mutually exclusive types: (1) {Entity Fragmentation}, where the model generates only a subset of tokens from the ground-truth entity; (2) {Data Asymmetry}, where the model simply repeats the queried subject entity instead of producing the correct target; and (3) {Relationship Sparsity}, where the output fails to include either the query entity or the correct target. 

\begin{figure}[t!]  
    \centering
    \includegraphics[width=\linewidth]{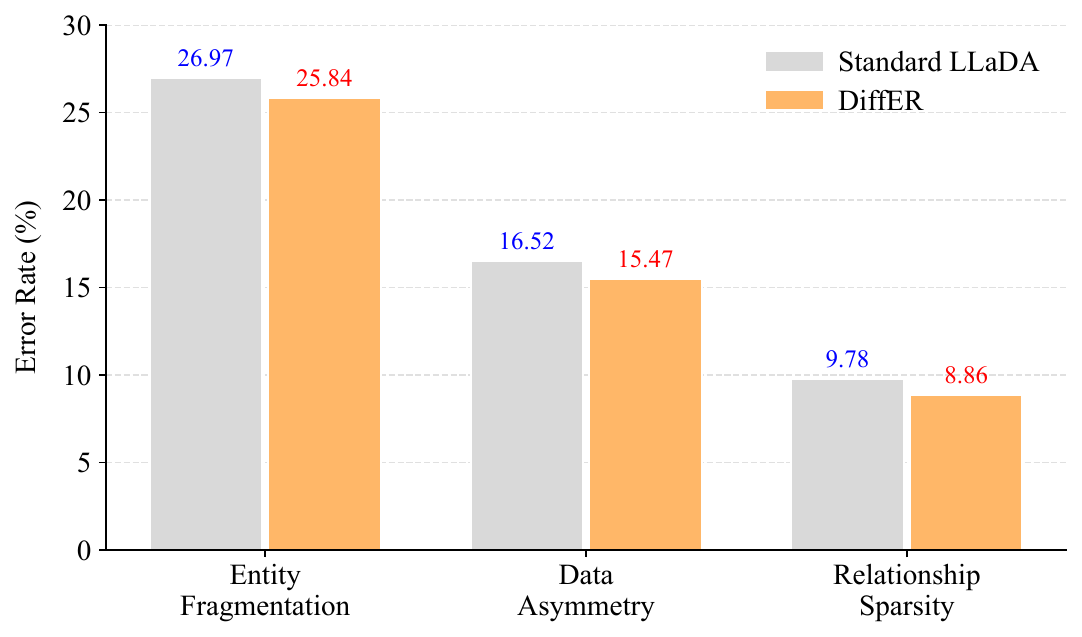}
    
    \caption{Error-type breakdown of inference failures under a unified evaluation setting. 
    }
    \label{tab:error_analysis}
\end{figure}

The results summarized in Fig.~\ref{tab:error_analysis} show that integrating DiffER leads to a substantial reduction in errors across all three categories compared with the standard LLaDA baseline. In particular, DiffER significantly decreases entity fragmentation errors, indicating a more robust capacity for holistic entity modeling. Errors arising from data asymmetry are also largely mitigated, suggesting that the model has developed stronger bidirectional reasoning ability. Finally, the reduction in relationship sparsity errors highlights an improved capability for logical and compositional inference. Overall, this analysis indicates that DiffER improves not only absolute factual accuracy but also the structural consistency of relational reasoning within the diffusion-generation paradigm.

\subsection{Case Study}
\begin{figure}[t!]  
    \centering
    \includegraphics[width=\linewidth]{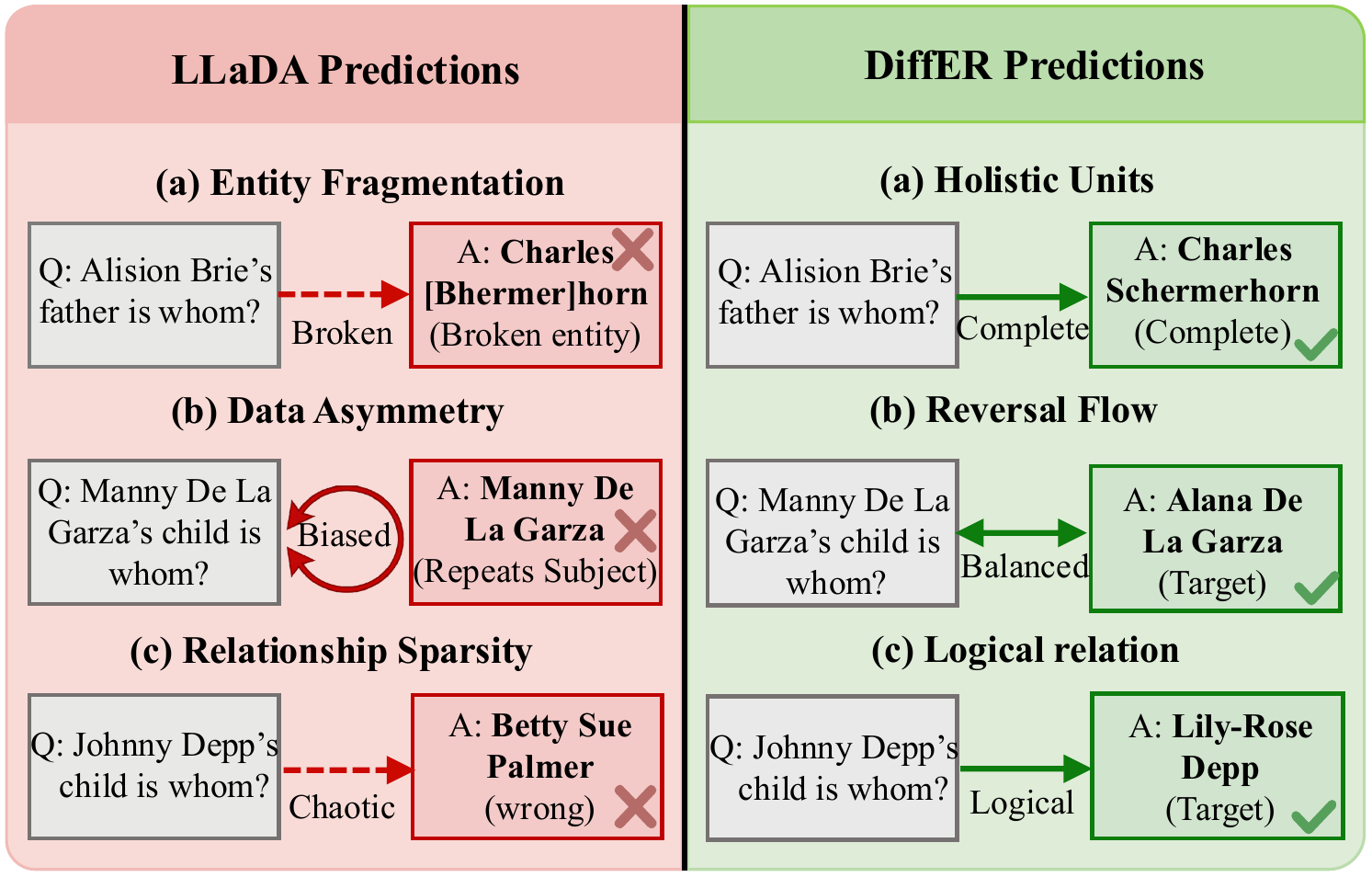}
    \caption{Typical predictions from LLaDA and DiffER. 
    }
    \label{tab:case_study}
\end{figure}

Fig.~\ref{tab:case_study} presents qualitative comparisons between DiffER and the LLaDA baseline, illustrating how DiffER addresses the failure modes identified in the error analysis. The baseline frequently fails to reverse learned relations and instead repeats the query subject. For example, 
Diffusion-based LLMs may generate corrupted multi-token entities due to iterative denoising. In our case, the baseline distorts ``Charles Schermerhorn'' into ``Charles Bhermerhorn,'' while DiffER produces the correct entity, demonstrating stronger entity-level coherence.
In addition, when asked ``Manny De La Garza’s child,'' it incorrectly outputs ``Manny De La Garza,'' whereas DiffER correctly predicts ``Alana De La Garza,'' indicating improved bidirectional relational reasoning.
Finally, the baseline often confuses inverse relations. For the query ``Johnny Depp’s child,'' it outputs an unrelated entity (``Betty Sue Palmer''), whereas DiffER correctly predicts ``Lily-Rose Depp,'' reflecting more accurate inverse relation modeling.





\section{Conclusion}
In this work, we show that Diffusion LLMs (DLLMs) remain susceptible to the reversal curse. Through a systematic analysis, we identify three fundamental causes: (1) \textit{Entity Fragmentation} in representations, (2) \textit{Data Asymmetry} in training distributions, and (3) \textit{Relation Sparsity} in logical reasoning. To address these issues, we propose the DiffER framework. It integrates (1) \textit{Whole-Entity Masking} to preserve entity-level semantic integrity, (2) \textit{Symmetric Alignment} to rebalance directional biases in the training data, and (3) \textit{Inverse Relation Modeling} to explicitly capture logical compositionality. Extensive experiments across multiple benchmarks show that DiffER substantially alleviates reversal curse failures, improving both forward and reverse reasoning. These results indicate that overcoming the reversal curse requires a coordinated approach addressing structural consistency and relational logic, offering a practical paradigm for enhancing reasoning in DLLMs.

\section*{Limitations}
While our proposed framework significantly mitigates the reversal curse in DLLMs, we acknowledge several limitations that warrant future investigation:
(1) Scalability of Data Construction: Our Symmetric Alignment and Inverse Relation Modeling strategies currently rely on structured relational triplets (e.g., Knowledge Graphs) to synthesize reverse data. Applying these strategies to unstructured, massive-scale web corpora remains a challenge, as automatically extracting and reversing complex, implicit relationships in natural text is non-trivial.
(2) Model Scale and Architecture: Our empirical validation was conducted primarily on the LLaDA-8B and Dream-7B architecture. While we believe the identified mechanisms (fragmentation, asymmetry, sparsity) are fundamental to diffusion models, further research is needed to verify whether scaling laws or different discrete diffusion architectures (e.g., MDLM) exhibit different behaviors regarding the reversal curse.

\section*{Ethics Statement}
This paper studies LLMs using publicly available pretrained models and benchmark datasets from open internet sources. The datasets employed in this work are widely adopted in the research community and contain no private user data or personally identifiable information. All models are evaluated as-is, without any additional training or fine-tuning that could amplify harmful behaviors. We therefore believe that this study complies with the ACL Ethics Policy.

\bibliography{custom}

\begin{thebibliography}{22}
\providecommand{\natexlab}[1]{#1}

\bibitem[{Allen-Zhu and Li(2023)}]{allen2023physics}
Zeyuan Allen-Zhu and Yuanzhi Li. 2023.
\newblock Physics of language models: Part 3.1, knowledge storage and extraction.
\newblock \emph{arXiv preprint arXiv:2309.14316}.

\bibitem[{Austin et~al.(2021)Austin, Johnson, Ho, Tarlow, and Van Den~Berg}]{austin2021structured}
Jacob Austin, Daniel~D Johnson, Jonathan Ho, Daniel Tarlow, and Rianne Van Den~Berg. 2021.
\newblock Structured denoising diffusion models in discrete state-spaces.
\newblock \emph{Advances in neural information processing systems}, 34:17981--17993.

\bibitem[{Berglund et~al.(2023)Berglund, Tong, Kaufmann, Balesni, Stickland, Korbak, and Evans}]{berglund2023reversal}
Lukas Berglund, Meg Tong, Max Kaufmann, Mikita Balesni, Asa~Cooper Stickland, Tomasz Korbak, and Owain Evans. 2023.
\newblock The reversal curse: Llms trained on" a is b" fail to learn" b is a".
\newblock \emph{arXiv preprint arXiv:2309.12288}.

\bibitem[{Cohen et~al.(2024)Cohen, Biran, Yoran, Globerson, and Geva}]{cohen2024evaluating}
Roi Cohen, Eden Biran, Ori Yoran, Amir Globerson, and Mor Geva. 2024.
\newblock Evaluating the ripple effects of knowledge editing in language models.
\newblock \emph{Transactions of the Association for Computational Linguistics}, 12:283--298.

\bibitem[{Cui et~al.(2021)Cui, Che, Liu, Qin, and Yang}]{cui2021pre}
Yiming Cui, Wanxiang Che, Ting Liu, Bing Qin, and Ziqing Yang. 2021.
\newblock Pre-training with whole word masking for chinese bert.
\newblock \emph{IEEE/ACM Transactions on Audio, Speech, and Language Processing}, 29:3504--3514.

\bibitem[{Dai et~al.(2022)Dai, Dong, Hao, Sui, Chang, and Wei}]{dai2022knowledge}
Damai Dai, Li~Dong, Yaru Hao, Zhifang Sui, Baobao Chang, and Furu Wei. 2022.
\newblock Knowledge neurons in pretrained transformers.
\newblock In \emph{Proceedings of the 60th Annual Meeting of the Association for Computational Linguistics (Volume 1: Long Papers)}, pages 8493--8502.

\bibitem[{Elazar et~al.(2021)Elazar, Kassner, Ravfogel, Ravichander, Hovy, Sch{\"u}tze, and Goldberg}]{elazar2021measuring}
Yanai Elazar, Nora Kassner, Shauli Ravfogel, Abhilasha Ravichander, Eduard Hovy, Hinrich Sch{\"u}tze, and Yoav Goldberg. 2021.
\newblock Measuring and improving consistency in pretrained language models.
\newblock \emph{Transactions of the Association for Computational Linguistics}, 9:1012--1031.

\bibitem[{Geva et~al.(2021)Geva, Schuster, Berant, and Levy}]{geva2021transformer}
Mor Geva, Roei Schuster, Jonathan Berant, and Omer Levy. 2021.
\newblock Transformer feed-forward layers are key-value memories.
\newblock In \emph{Proceedings of the 2021 Conference on Empirical Methods in Natural Language Processing}, pages 5484--5495.

\bibitem[{Gong et~al.(2022)Gong, Li, Feng, Wu, and Kong}]{gong2022diffuseq}
Shansan Gong, Mukai Li, Jiangtao Feng, Zhiyong Wu, and LingPeng Kong. 2022.
\newblock Diffuseq: Sequence to sequence text generation with diffusion models.
\newblock \emph{arXiv preprint arXiv:2210.08933}.

\bibitem[{Grosse et~al.(2023)Grosse, Bae, Anil, Elhage, Tamkin, Tajdini, Steiner, Li, Durmus, Perez et~al.}]{grosse2023studying}
Roger Grosse, Juhan Bae, Cem Anil, Nelson Elhage, Alex Tamkin, Amirhossein Tajdini, Benoit Steiner, Dustin Li, Esin Durmus, Ethan Perez, and 1 others. 2023.
\newblock Studying large language model generalization with influence functions.
\newblock \emph{arXiv preprint arXiv:2308.03296}.

\bibitem[{He et~al.(2023)He, Sun, Tang, Wang, Huang, and Qiu}]{he2023diffusionbert}
Zhengfu He, Tianxiang Sun, Qiong Tang, Kuanning Wang, Xuan-Jing Huang, and Xipeng Qiu. 2023.
\newblock Diffusionbert: Improving generative masked language models with diffusion models.
\newblock In \emph{Proceedings of the 61st annual meeting of the association for computational linguistics (volume 1: Long papers)}, pages 4521--4534.

\bibitem[{Li et~al.(2022)Li, Thickstun, Gulrajani, Liang, and Hashimoto}]{li2022diffusion}
Xiang Li, John Thickstun, Ishaan Gulrajani, Percy~S Liang, and Tatsunori~B Hashimoto. 2022.
\newblock Diffusion-lm improves controllable text generation.
\newblock \emph{Advances in neural information processing systems}, 35:4328--4343.

\bibitem[{Lou et~al.(2023)Lou, Meng, and Ermon}]{lou2023discrete}
Aaron Lou, Chenlin Meng, and Stefano Ermon. 2023.
\newblock Discrete diffusion modeling by estimating the ratios of the data distribution.
\newblock \emph{arXiv preprint arXiv:2310.16834}.

\bibitem[{Lu et~al.(2024)Lu, Jin, Li, Tian, Zhang, Wang, Xu, Tian, and Cai}]{lu2024rethinking}
Zhicong Lu, Li~Jin, Peiguang Li, Yu~Tian, Linhao Zhang, Sirui Wang, Guangluan Xu, Changyuan Tian, and Xunliang Cai. 2024.
\newblock Rethinking the reversal curse of llms: a prescription from human knowledge reversal.
\newblock In \emph{Proceedings of the 2024 Conference on Empirical Methods in Natural Language Processing}, pages 7518--7530.

\bibitem[{Ma et~al.(2023)Ma, Gu, Ling, Liu, and Liu}]{ma2023untying}
Jun-Yu Ma, Jia-Chen Gu, Zhen-Hua Ling, Quan Liu, and Cong Liu. 2023.
\newblock Untying the reversal curse via bidirectional language model editing.
\newblock \emph{arXiv preprint arXiv:2310.10322}.

\bibitem[{Meng et~al.(2022)Meng, Bau, Andonian, and Belinkov}]{meng2022locating}
Kevin Meng, David Bau, Alex Andonian, and Yonatan Belinkov. 2022.
\newblock Locating and editing factual associations in gpt.
\newblock \emph{Advances in neural information processing systems}, 35:17359--17372.

\bibitem[{Nie et~al.(2025)Nie, Zhu, You, Zhang, Ou, Hu, Zhou, Lin, Wen, and Li}]{nie2025large}
Shen Nie, Fengqi Zhu, Zebin You, Xiaolu Zhang, Jingyang Ou, Jun Hu, Jun Zhou, Yankai Lin, Ji-Rong Wen, and Chongxuan Li. 2025.
\newblock Large language diffusion models.
\newblock \emph{arXiv preprint arXiv:2502.09992}.

\bibitem[{Pan et~al.(2025)Pan, Hahami, Fan, Xie, and Sompolinsky}]{pan2025closing}
Xu~Pan, Ely Hahami, Jingxuan Fan, Ziqian Xie, and Haim Sompolinsky. 2025.
\newblock Closing the data-efficiency gap between autoregressive and masked diffusion llms.
\newblock \emph{arXiv preprint arXiv:2510.09885}.

\bibitem[{Sahoo et~al.(2024)Sahoo, Arriola, Schiff, Gokaslan, Marroquin, Chiu, Rush, and Kuleshov}]{sahoo2024simple}
Subham Sahoo, Marianne Arriola, Yair Schiff, Aaron Gokaslan, Edgar Marroquin, Justin Chiu, Alexander Rush, and Volodymyr Kuleshov. 2024.
\newblock Simple and effective masked diffusion language models.
\newblock \emph{Advances in Neural Information Processing Systems}, 37:130136--130184.

\bibitem[{Shi et~al.(2024)Shi, Han, Wang, Doucet, and Titsias}]{shi2024simplified}
Jiaxin Shi, Kehang Han, Zhe Wang, Arnaud Doucet, and Michalis Titsias. 2024.
\newblock Simplified and generalized masked diffusion for discrete data.
\newblock \emph{Advances in neural information processing systems}, 37:103131--103167.

\bibitem[{Ye et~al.(2025)Ye, Xie, Zheng, Gao, Wu, Jiang, Li, and Kong}]{ye2025dream}
Jiacheng Ye, Zhihui Xie, Lin Zheng, Jiahui Gao, Zirui Wu, Xin Jiang, Zhenguo Li, and Lingpeng Kong. 2025.
\newblock Dream 7b: Diffusion large language models.
\newblock \emph{arXiv preprint arXiv:2508.15487}.

\bibitem[{Zhu et~al.(2024)Zhu, Huang, Zhang, Jordan, Jiao, Tian, and Russell}]{zhu2024towards}
Hanlin Zhu, Baihe Huang, Shaolun Zhang, Michael Jordan, Jiantao Jiao, Yuandong Tian, and Stuart~J Russell. 2024.
\newblock Towards a theoretical understanding of the'reversal curse'via training dynamics.
\newblock \emph{Advances in Neural Information Processing Systems}, 37:90473--90513.

\end{thebibliography}

\end{document}